\documentclass[lettersize,journal]{IEEEtran}
\usepackage{amsmath,amsfonts}
\usepackage{algorithmic}
\usepackage{algorithm}
\usepackage{array}
\usepackage[caption=false,font=footnotesize]{subfig}
\usepackage{stfloats}
\usepackage{url}
\usepackage{verbatim}
\usepackage{graphicx}
\usepackage{amsmath}
\usepackage{mathrsfs}  
\usepackage{threeparttable}
\def\BibTeX{{\rm B\kern-.05em{\sc i\kern-.025em b}\kern-.08em
    T\kern-.1667em\lower.7ex\hbox{E}\kern-.125emX}}
\usepackage{amssymb}
\usepackage{color}
\usepackage{colortbl}
\usepackage{tabulary}
\definecolor{Gray}{gray}{0.9}
\usepackage{amsmath}
\usepackage{subfloat}
\usepackage[utf8]{inputenc} 
\usepackage[T1]{fontenc}    
\usepackage{url}            
\usepackage{booktabs}       
\usepackage{amsfonts}       
\usepackage{nicefrac}       
\usepackage{microtype}      
\usepackage{xcolor}         
\usepackage{anyfontsize}
\usepackage{multirow}
\usepackage{wrapfig}
\usepackage{hyperref}
\usepackage{url}
\usepackage{arydshln}
\usepackage[numbers,sort]{natbib}
\usepackage{enumitem}
\usepackage{amsthm,amssymb}
\usepackage{algorithm}
\usepackage{graphicx}
\usepackage{colortbl}
\usepackage{dsfont}
\usepackage{float} 
\usepackage{makecell}
\usepackage{diagbox}
\usepackage{fancybox}
\usepackage{bm}

\hypersetup{hidelinks,
	colorlinks=true,
	linkcolor=black,
        filecolor=black,      
        urlcolor=blue,
	pdfstartview=Fit,
        citecolor=blue,
	breaklinks=true
}

\begin{document}

\title{Debunk and Infer: Multimodal Fake News Detection via Diffusion-Generated Evidence and LLM Reasoning}

\author{Kaiying Yan,
        Moyang Liu,
        Yukun Liu,
        Ruibo Fu,
        Zhengqi Wen,
        Jianhua Tao,
        and Xuefei Liu, ~\IEEEmembership{Fellow,~IEEE}

\thanks{Kaiying Yan is with Sun Yat-sen University, Guangzhou, Guangdong, 510275, China (e-mail: yanky6@mail2.sysu.edu.cn).}
\thanks{Moyang Liu is with Beihang University, Beijing, 100191, China (e-mail: moyang\_liu@buaa.edu.cn).}
\thanks{Yukun Liu is with University of Chinese Academy of Sciences, Beijing, 100049, China (e-mail: yukunliu927@gmail.com).}
\thanks{Ruibo Fu and Xuefei Liu are with Institute of Automation, Chinese Academy of Sciences, Beijing, 100190, China (e-mail: \{ruibo.fu, xufei.liu\}@nlpr.ia.ac.cn).}
\thanks{Zhengqi Wen is with Beijing National Research Center for Information Science and Technology, Tsinghua University, Beijing, 100084, China (e-mail: zqwen@tsinghua.edu.cn).}
\thanks{Jianhua Tao is with Department of Automation, Tsinghua University, Beijing, 100084, China, and also Beijing National Research Center for Information Science and Technology, Tsinghua University, Beijing, 100084, China (e-mail: jhtao@tsinghua.edu.cn).}

\thanks{This work is supported by National Natural Science Foundation of China (NSFC) (No.62306316, No.62101553, No.U21B20210).
}

}



\maketitle

\begin{abstract}
The rapid spread of fake news across multimedia platforms presents serious challenges to information credibility. In this paper, we propose a Debunk-and-Infer framework for Fake News Detection(DIFND) that leverages debunking knowledge to enhance both the performance and interpretability of fake news detection. DIFND integrates the generative strength of conditional diffusion models with the collaborative reasoning capabilities of multimodal large language models (MLLMs). Specifically, debunk diffusion is employed to generate refuting or authenticating evidence based on the multimodal content of news videos, enriching the evaluation process with diverse yet semantically aligned synthetic samples. To improve inference, we propose a chain-of-debunk strategy where a multi-agent MLLM system produces logic-grounded, multimodal-aware reasoning content and final veracity judgment. By jointly modeling multimodal features, generative debunking cues, and reasoning-rich verification within a unified architecture, DIFND achieves notable improvements in detection accuracy. Extensive experiments on the FakeSV and FVC datasets show that DIFND not only outperforms existing approaches but also delivers trustworthy decisions.
\end{abstract}


\section{Introduction}
\IEEEPARstart{I}n recent years, the rapid proliferation of misinformation has posed significant threats to public discourse, political stability, and public health. 
The rise of multimedia platforms like YouTube and TikTok\cite{jordan2024rise} has fueled the rapid spread of fake news videos, making it increasingly difficult to separate truth from falsehood. To address this challenge, automated fake news detection models can enhance screening efficiency and help maintain a truthful online environment.

Early unimodal approaches focused on extracting statistical features from text\cite{shu2019beyond, serrano2020nlp} or modeling temporal consistency\cite{yang2019exposing} for classification. However, relying solely on single-modal methods makes it challenging to detect multimodal fake news effectively. With the advancement of relevant technologies, multimodal detection has become increasingly important. This led to the emergence of dual-modal fusion methods, primarily focusing on interactions between textual and visual modalities\cite{singhal2020spotfake+, wang2022fmfn, peng2024not}. These methods explored inter- and intra-modal relationships\cite{wei2023modeling}, contrastive learning techniques\cite{wang2023cross}, and other strategies. With further progress in deep learning, modern fake news detectors have begun to incorporate additional modalities such as audio and video\cite{qi2023fakesv, song2021multimodal}. Unlike traditional text-based or text-visual fake news, multimodal news videos often exploit coherence across subtitles, visuals, speech, background music, and more-making them more persuasive and significantly harder to detect using unimodal analysis alone.

Recent advances have begun to explore fine-grained modeling of debunking, highlighting its critical role in fake news detection. Debunking serves as a corrective mechanism by providing verified evidence to refute false claims, thereby enhancing the interpretability and trustworthiness of detection systems. Several studies have investigated the integration of debunking into fake news detection pipelines, such as constructing datasets to capture spontaneous user debunking behavior\cite{miyazaki2023fake}, modeling multimodal consistency between suspicious news content and existing debunking materials related to the same events\cite{qi2023two}, and applying multiple instance learning (MIL) frameworks to assess the credibility of individual sentences\cite{yang2023wsdms}.
Despite these promising developments, debunking data continues to pose significant challenges. It typically depends on manually curated fact-checks or counter-narratives, which are limited in both quantity and scope\cite{glockner2022missing}, restricting the scalability and generalizability of learning-based models. Existing datasets such as AMBIFC\cite{glockner2024ambifc} and CREDULE\cite{chrysidis2024credible} offer valuable resources, but they are primarily composed of textual and image content derived from web articles, rendering them less suitable for short-form video debunking scenarios. As a result, the acquisition, enhancement, and utilization of debunking data for fake news video detection remain largely unexplored and warrant further research.

In the era of large-scale models, as emphasized in recent surveys\cite{zhou2024survey, chai2025text, cheng2025empowering}, multimodal large language models (MLLMs) have demonstrated remarkable capabilities in both generation and reasoning. These models have emerged as valuable tools in advancing multimodal fake news detection. For example, LVLMs have been enhanced with manipulated media containing fake information\cite{liu2024fka}, and prior knowledge has been incorporated into LVLMs to improve performance in misinformation scenarios\cite{liu2024fakenewsgpt4}.
Despite these advancements, several challenges remain. Key concerns include the potential introduction of artifacts or biases during the data augmentation process\cite{chalehchaleh2024enhancing}, limited reasoning capabilities in complex misinformation contexts\cite{hu2024bad}, and a general lack of interpretability in model outputs\cite{wang2024explainable}. Moreover, the built-in safeguards of LLMs often restrict the generation of diverse misinformation-related content, resulting in minimal and inconsistent benefits in downstream fake news detection tasks\cite{choi2025limited}.

In this work, we propose a novel Debunk-and-Infer framework for multimodal fake news detection, consisting of debunk diffusion and a chain-of-debunk strategy within a multi-agent MLLM system. 
We begin by broadening the definition of debunking to encompass both the refutation of fake claims and the verification of true ones, thereby promoting data balance and improving model training. The proposed debunk diffusion approach generates context-aware features conditioned on multimodal news content, enabling the model to produce debunking representations that are semantically consistent with the supporting evidence. To address the scarcity of debunking data, we augment the dataset with LLM-generated content based
on verified annotations, including rationales and contextual cues. The chain-of-debunk strategy decomposes the verification process into a sequence of structured reasoning steps, enabling collaborative inference among multiple MLLMs. Finally, we employ attention-based fusion modules and decision-level aggregation to effectively integrate debunking cues and multimodal features, ultimately enhancing the accuracy and robustness of the final decision.

The main contribution of this paper is:
\begin{enumerate}
    \item  We propose DIFND, a debunk-and-infer framework that integrates debunking knowledge with collaborative inference to enhance multimodal fake news detection. By incorporating diverse generated evidence from the proposed debunk diffusion and chain-of-debunk from multi-agent MLLMs, the framework facilitates more accurate and trustworthy misinformation identification.
    \item We develop Debunk Diffusion, a mechanism for generating textual features enriched with debunking cues. Trained on an LLM-augmented debunking dataset, a conditional latent diffusion model leverages multimodal content to synthesize high-quality debunking representations, which enhance the model’s capacity to recognize and reason about news authenticity more effectively.
    \item Extensive experiments on the FakeSV and FVC datasets demonstrate that our proposed DIFND framework outperforms existing methods, as it adaptively fuses multimodal features, plausible evidence generated by debunk diffusion, and reasoning-rich assessments derived from MLLMs.
\end{enumerate}

\begin{figure*}[!t]
    \vspace{-2em}
	\centering
	\includegraphics[width=0.97\textwidth]{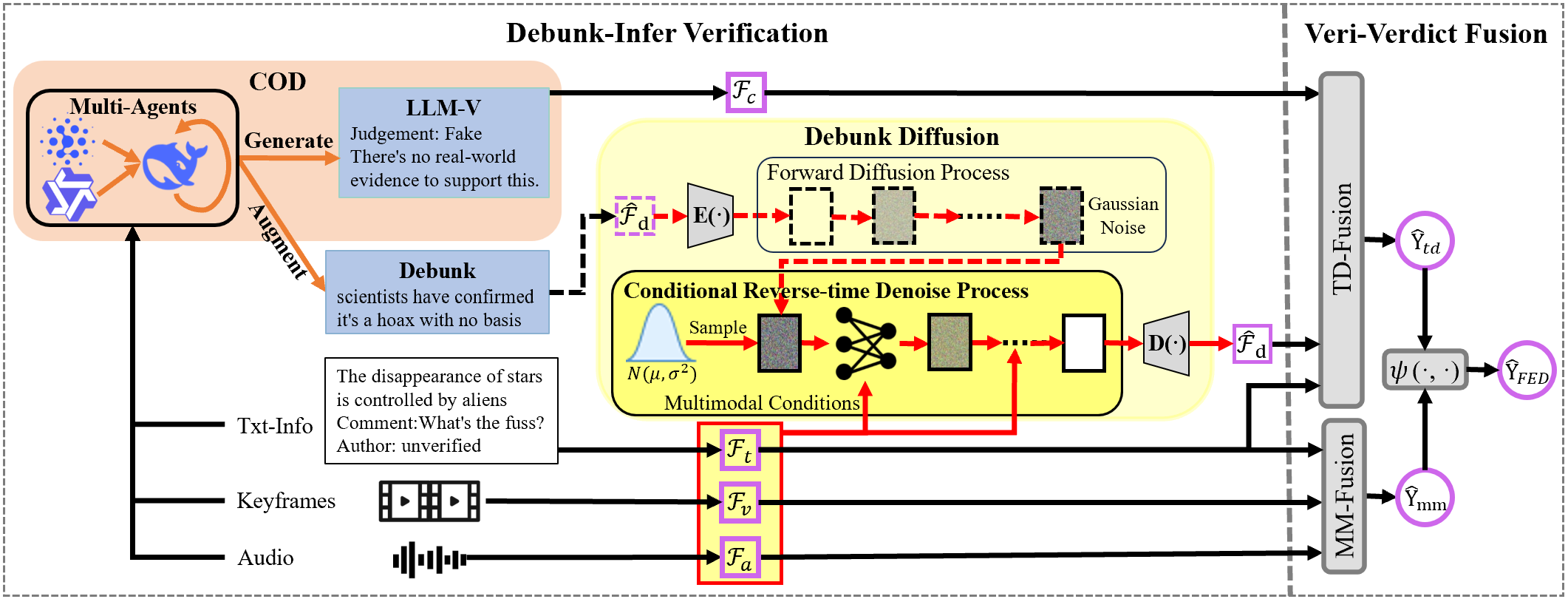}
 \vspace{-1em}
	\caption{Overview of DIFND Framework. The DIFND framework consists of Debunk-Infer Verification and Veri-Verdict Fusion. Debunk Diffusion generates debunking cues conditioned on multimodal inputs and is trained on the LLM-augmented dataset, while multi-agent LLMs perform chain-of-debunk for reasoning-rich verification named LLM-V. Final decisions are made via attentive fusion of features from all modules. The dashed arrows indicate paths that are used only during training and are not involved during inference. The textual information with blue background is generated or enhanced by LLMs.}
 \label{fig:pipeline}
 \vspace{-1.5em}
\end{figure*}

\section{Related Work}

\subsection{Multimodal Fake News Detection}
Recent advancements in multimodal fake news detection have focused on enhancing cross-modal reasoning, integrating external knowledge, and incorporating MLLMs into the verification process. FakingRecipe\cite{bu2024fakingrecipe} approaches the issue from the perspective of fake news video creation, analyzing material selection and editing to capture traces of tampering. MIMoE-FND\cite{liu2025modality} employs a hierarchical mixture-of-experts approach with interaction gating to model complex modality interactions, capturing both semantic alignment and prediction agreement. DSEN-EK\cite{qiu2025dsen} strengthens entity relationship modeling and the representation of important entities through a two-layer semantic information extraction network and an external knowledge integration network. To address data scarcity, CMA\cite{jiang2025cross} transforms few-shot learning into a robust zero-shot problem via cross-modal augmentation, enhancing performance with minimal labeled data.

Moreover, the integration of multimodal large language (MLLMs) into the fake news detection pipeline has emerged as a promising direction. FKA-Owl \cite{liu2024fka} enhances vision-language models (LVLMs) with forgery-specific knowledge, enabling them to reason more effectively about manipulation details across both textual and visual modalities. Similarly, FakeNewsGPT4 \cite{liu2024fakenewsgpt4} incorporates knowledge graphs and forgery-related priors into LVLMs to improve generalization and cross-domain robustness. These efforts collectively push the boundaries of multimodal fake news detection by addressing key challenges such as modality fusion, knowledge integration, and model scalability. In another line of work, LESS4FD \cite{ma2024fake} utilizes LLMs to construct a heterogeneous graph that captures complex relationships among news articles, entities, and topics. It further employs a generalized PageRank algorithm guided by a consistency learning objective to extract both local and global semantic signals.

\subsection{Diffusion Model}
Diffusion models, originally developed for continuous data domains such as images and audio, have recently been adapted for natural language processing (NLP) tasks, including text data augmentation and generation\cite{yi2024diffusion}.

In the realm of text generation, several studies have explored the application of diffusion models. 
DiffuSeq\cite{gong2022diffuseq} introduces a sequence-to-sequence framework that leverages diffusion processes to generate text. 
Lovelace et al.\cite{lovelace2023latent} combined the pre-trained encoder-decoder and the diffusion model, thus transforming the text generation problem into a continuous diffusion process in language latent space. 
Li et al. developed Diffusion-LM\cite{zhou2024difflm}, a non-autoregressive language model based on continuous diffusion, which enables controllable generation of text.

In terms of data augmentation, diffusion models have been utilized to generate synthetic text data to improve model performance in low-resource settings.
DiffusionCLS\cite{chen2024effective} employs a diffusion language model to generate pseudo-samples by reconstructing sentiment-bearing tokens, thereby enhancing sentiment classification tasks in scenarios with limited data.
Wang et al.\cite{wang2023incomplete} use a score-based diffusion model to recover data from missing modalities and guide emotion recognition in the presence of incomplete multimodal data by conditioning on the available modalities.
Nguyen et al.\cite{blow2025data} proposed a data augmentation method based on a diffusion model, aiming to improve the fairness of AI systems by generating synthetic data.

\subsection{Large Language Model}
Recent advancements in inference for large language models (LLMs) have introduced several innovative methodologies aimed at enhancing the models' reasoning capabilities.
Chain-of-thought prompting \cite{wei2022chain} significantly improves the performance of large language models on reasoning tasks by including a series of examples of intermediate reasoning steps in the prompt. ReAct \cite{yao2023react} synergizes reasoning and acting in large language models by prompting them to generate both reasoning traces and task-specific actions in an interleaved manner. Chen et al.\cite{chen2025lvagent} proposed LVAgent, a framework for long video understanding that enables multi-round dynamic collaboration among multimodal large language model (MLLM) agents. \cite{shi2024unlocking} proposed Agent-of-Thoughts Distillation (AoTD), using an agent-based system to decompose complex questions into sub-tasks and specialized vision models to generate reasoning chains which are used to conduct knowledge distillation.

Several recent studies have explored the use of large language models (LLMs) for reasoning in fake news detection. Liu et al.\cite{liu2024detect} proposed the Dual-perspective Knowledge-guided Fake News Detection (DKFND) model, which enhances LLMs in few-shot settings by incorporating both internal and external knowledge for improved inference. Hu et al.\cite{hu2024bad} introduced an Adaptive Justification Guidance (ARG) network that boosts the performance of small language models (SLMs) by selectively leveraging justifications generated by LLMs and integrating knowledge distillation techniques. Zhang et al.\cite{zhang2025llms} developed a novel approach that generates both supportive and contradictory reasoning using LLMs, and trains models to learn semantic consistency between the news content and the generated reasoning.

\section{Methods}
\subsection{Overview}

The overview framework is illustrated in Figure~\ref{fig:pipeline}, which comprises two processes: Debunk-Infer Verification and Veri-Verdict Fusion, where Debunk-Infer Verification consists of Debunk Diffusion and Chain of Debunk.

News videos encompass a diverse range of multimodal information, incorporating elements such as text, audio, keyframes, debunking text, comments, and associated publisher profiles.
Multimodal features are obtained by specific unimodal feature extractors, while text modality is extracted by two additional multimodal-aligned feature extractors for audio and vision alignment.
These features will be interactively integrated and used for Debunk-Infer Verification and the VeriFusion process.

For Debunk-Infer Verification, the debunk diffusion module is a key component of the proposed framework, designed to model the distribution of latent debunking features. During the denoising process, these features are sampled from the learned distribution, conditioned on multimodal content to guide the generation. This mechanism enables the framework to support news video authentication even in scenarios where explicit debunking information is unavailable for a given news item.
It is important to note that in many real-world datasets, debunking text may be sparse or entirely absent for individual samples, despite its critical role in the overall pipeline. To address this limitation and enrich the debunking dataset, we employ tailored prompt engineering for large language models (LLMs) to generate and expand debunking text, thereby enhancing both data diversity and semantic depth.

To enhance the interpretability of misinformation detection, we propose a chain-of-debunk (COD) strategy that employs multiple multimodal large language (MLLMs) as collaborative agents within a multi-stage reasoning framework. These agents are prompted to generate reason-rich verification named LLM-V through a progressive process that begins with unimodal analysis such as textual credibility evaluation, followed by the interpretation of visual and auditory cues through video and audio captioning. The outputs from each modality are then synthesized into a comprehensive, cross-modal reasoning assessment. This collaborative inference process enables the system to produce more transparent and explainable verification outcomes by explicitly articulating supporting or refuting evidence across different modalities.

The features extracted or generated from the multimodal information, debunking diffusion, and COD modules are fused using attention mechanisms to obtain a refined representation for the final decision.


\subsection{Debunk Diffusion}\label{Debunk Diffusion}

To begin with, we expand the conventional definition of debunking, which is typically limited to the refutation of false claims. Instead, we propose a more comprehensive and inclusive conceptualization that encompasses not only the disproving of misinformation but also the authentication of truthful information. By broadening this scope, our approach recognizes the dual necessity of both rejecting falsehoods and verifying facts, thereby providing a more balanced and holistic framework as well as a corresponding augmented dataset for evaluating the credibility of news videos.

As noted in the Introduction, the potential of debunking data remains underexplored, and such data is often scarce or entirely absent in many real-world datasets. To address this limitation, we design a debunk diffusion network that models the distribution of latent debunking features conditioned on the multimodal information of news videos. By sampling from this learned distribution, the network generates informative debunking cues, enabling more reliable news authentication even when explicit debunking data is unavailable for a given news item.

\begin{figure}[H]
        \vspace{-1em}
	\centering
	\includegraphics[width=0.5\textwidth]{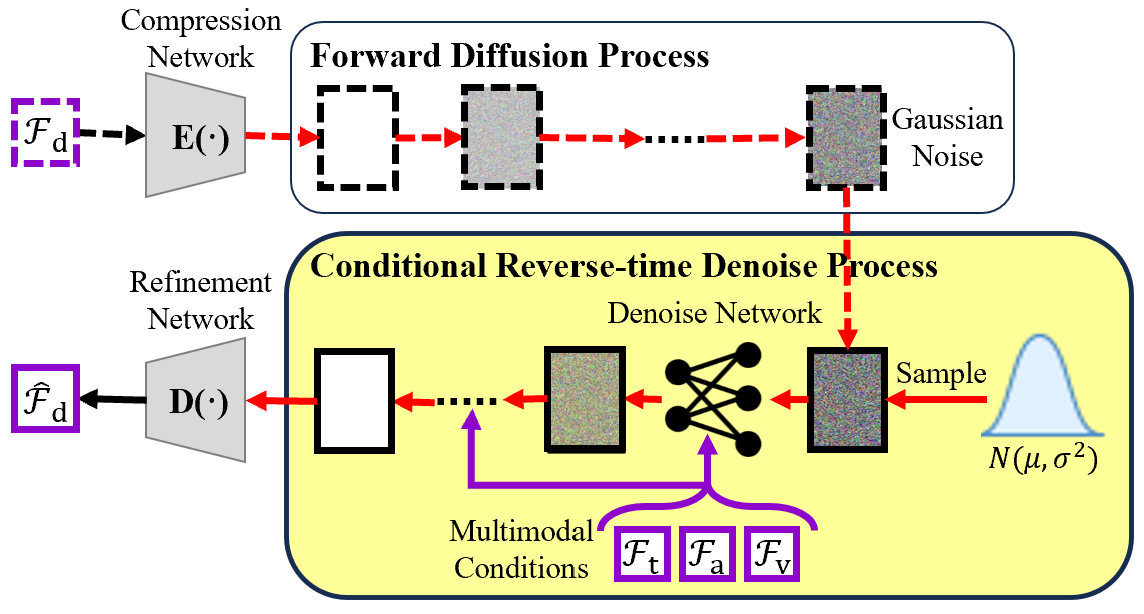}
 \vspace{-1.6em}
	\caption{Overview of conditional latent debunk diffusion.}
 \label{fig:denoise}
 \vspace{-0.8em}
\end{figure}

Figure~\ref{fig:denoise} provides an overview of latent debunk diffusion, which primarily consists of four key components: compression network, conditional latent diffusion, multimodal conditioning mechanism, and refinement network. First, we employ a learnable compression network to transform the original high-dimensional text features into a compact latent space. Next, a continuous conditional diffusion model is trained to generate samples from the latent distribution of our language compressor, guided by the multimodal conditioning mechanism. Finally, generated latent features are self-refined, which are regarded as the sampled debunk latent features and will be used in subsequent feature fusion.

\subsubsection{Compression Network} The learnable compression network transforms the encoded high-dimensional text features into a compact latent space optimized for diffusion, as key information of debunk text needs to be extracted and directly using the original long-sequence features would result in significant deviations. To achieve this, we adopt the Perceiver Resampler architecture and designed to efficiently compress high-dimensional data, such as image or text features, into a lower-dimensional latent representation.
The language encoder $\Theta(\cdot)$ transforms a variable-length language input, represented as a sequence of tokens $ \boldsymbol{w} \in \mathbb{N}^L$, into a latent representation $ \Theta(\boldsymbol{w}) \in \mathbb{R}^{L \times d_{LE}}$ of the same length, that is the debunk feature $\mathcal{F}_d$ mentioned above. The compression network, donated as $E(\cdot)$, utilizes $\ell$ learned latent queries $\mathcal{Q} \in\mathbb{R}^{\ell\times d}$ to attend both to themselves and to the fixed encoder representations through cross-attention simultaneously, compressing the encoder representations to the fixed sequence length $\ell$ and reduce the dimensionality from $d_{LE}$ to $d$ through linear projection, enabling efficient information compression while preserving essential details. 
So the compression process is written as
\begin{align}
    \boldsymbol{z} &= E(\Theta(\mathbf{w}))= E(\mathcal{F}_d) \in \mathbb{R}^{\ell \times d} \\
    &=\mathrm{FF} (\mathrm{FF}(\mathcal{Q})+ \mathrm{MHA} (q=\mathcal{Q},kv=[\mathcal{Q};\mathcal{F}_d]))
\end{align}
where $\mathrm{MHA}$,$\mathrm{FF}$ refer to the multi-head attention block, feedforward layers respectively and we fixed length $\ell < L$ and dimensionality $d < d_{LE}$ where we will learn our diffusion model. To train the model, we apply mean pooling to the output latent representations $\boldsymbol{z}$, aggregate them into a fixed-size vector passed to a classifier for prediction and get the loss $\mathcal{L}_d$ using cross-entropy.


\subsubsection{Conditional Latent Diffusion}
Diffusion models learn a series of state transitions to map noise from a known prior distribution to $x_0$ from the data distribution. To achieve this, they first define a forward transition from data to noise:
\begin{align}
    \boldsymbol{z}_t = \sqrt{\gamma(t)} \boldsymbol{z}_0 + \sqrt{1 - \gamma(t)} \epsilon
\end{align}
where $ \epsilon \sim \mathcal{N}(0, I) $, $ t \sim \mathcal{U}(0, T) $ is a continuous variable, and $ \gamma(t) $ is a monotonically decreasing function from 1 to 0.

Given a news debunk language dataset $ \hat{D} $, we sample continuous data $\boldsymbol{z}=g_{\phi}(\Theta(\mathbf{w}))\in \mathbb{R}^{\ell \times d}$, where $\mathbf{w} \sim \hat{D}$. Subsequently, we train a continuous denoising network $f(\boldsymbol{z}_t,\mathcal{X}_{cond},t)$ to reconstruct $ \boldsymbol{z} $ using a standard regression objective, where $\mathcal{X}_{cond}$ refers to the conditions consisting of the multimodal feature of news videos. This training of $f(\boldsymbol{z}_t,\mathcal{X}_{cond},t)$ is based on denoising with a MSE loss:
\begin{align}
    \mathcal{L}_{\text{mse}} = \underset{
          \substack{
            t \sim \mathcal{U}(0,T) \\ 
            \epsilon \sim \mathcal{N}(0,1)
          }
        }
    {\mathbb{E}}\left\| f\left(\sqrt{\gamma(t)} \, \boldsymbol{z}^0 + \sqrt{1 - \gamma(t)} \, \epsilon, \mathcal{X}_{cond},t\right) - \boldsymbol{z}^0 \right\|^2
\end{align}
where $\mathcal{X}_{cond}=[\mathcal{F}_t, \mathcal{F}_a, \mathcal{F}_v]$ denotes a fused feature consisting of multimodal features.

For each generation, we start from a latent variable $ \boldsymbol{z_s}=\boldsymbol{z}^T \in \mathbb{R}^{\ell \times d} \sim \mathcal{N}(0, I) $ and conduct a series of denoised transition $\boldsymbol{z}^{T} \longrightarrow\boldsymbol{z}^{T-1} \longrightarrow \cdots \longrightarrow \boldsymbol{z}^0$ to produce a sample $\tilde{\boldsymbol{z}}=\boldsymbol{z}^0$, from the distribution of the debunk latent space.
We can iteratively transition to $ \boldsymbol{z}^{t-\Delta} $ using the estimated $ \tilde{\boldsymbol{z}}^0 $ through repeatedly applying the denoising function $ f $ on each state $ \boldsymbol{z}^t $ to estimate $ \boldsymbol{z}^0 $. The above process can be accomplished by transition rules such as those specified in DDIM.

\subsubsection{Multimodal Conditioning Mechanism}
How to embed the multimodal conditions into the denoised process is a key question. We utilize learnable positional encoding and the cross-attention mechanism to aggregate multimodal features and encode them into the sample generation. The condition's position encoding is generated by the Sinusoidal position encoding for each modality sequence feature plus learnable modal offsets to distinguish different modalities. The multimodal sequence features added by position encoding are concatenated into conditions, donated as $\mathcal{X}_{cond}=[\mathcal{F}_t, \mathcal{F}_a, \mathcal{F}_v]$.
The conditioning mechanism can be defined as:

\begin{align}
\boldsymbol{z}_{cond}^t &= \text{softmax}\left(\frac{QK^\top}{\sqrt{d}}\right)V,\\
\boldsymbol{z}^t &= \boldsymbol{z}^t +\boldsymbol{z}_{cond}^t
\label{Eq: attn}
\end{align}

where $\boldsymbol{z}_{cond}^t$ is condition embedded representation, $Q = \boldsymbol{z}^t \mathbf{W}_Q$, $K = \mathcal{X}_{cond}\mathbf{W}_K$, $V = \mathcal{X}_{cond}\mathbf{W}_V$, and $\mathbf{W}_Q$,$\mathbf{W}_K$,$\mathbf{W}_V$
are the learnable parameters.
Our denoising network is a pre-LayerNorm multi-layer transformer, so the cross-attention mechanism Eq.~\ref{Eq: attn} is applied to each layer of the network to embed multimodal conditions.

\subsubsection{Self-Refinement Network}
Subsequently, to reduce noise in features, the denoised debunk features $\tilde{\boldsymbol{z}}$ are fed into the self-refinement network to obtain the refined debunk features $\hat{\mathcal{F}_d}=\hat{\mathcal{Z}} = \mathrm{D}(\tilde{\boldsymbol{z}})$, $\mathrm{D}(\cdot)$ denotes the self-refinement network of debunk features. The training objective if to minimize the reconstruction loss $\mathcal{L}_{\text{rec}}$ defined as:
\begin{align}
    \mathcal{L}_{\text{rec}} =  \| \hat{\mathcal{Z}} - \boldsymbol{z} \|_{2}^{2}.
\end{align}
Finally, to further guide the denoise objective, we apply mean pooling to the denoise latent features$\hat{\boldsymbol{z}}$ and then feed it into a classifier to obtain loss $\hat{\mathcal{L}}_d$. So the overall training objective is
\begin{align}
    \mathcal{L}_{\text{diff}} =  \mathcal{L}_{\text{mse}} + \mathcal{L}_{\text{rec}} + \mathcal{L}_{d} +\hat{\mathcal{L}}_d
\end{align}

\subsubsection{Debunk Text Augmentation by LLM}\label{debunk_aug}
In many cases, debunk texts are not consistently available across all samples in the dataset, which poses a challenge for training a robust model. To address this issue, we leverage large language models (LLMs) to perform data augmentation by synthesizing various debunk texts. Our approach integrates three sources of information: (1) the textual information of the news video itself, (2) related news covering the same event, and (3) existing debunk texts provided in the dataset. By prompting the LLM with this contextual information, we generate plausible debunking statements, even for samples lacking explicit refutations.

\begin{figure}[htbp]
	\centering
	\includegraphics[width=0.48\textwidth]{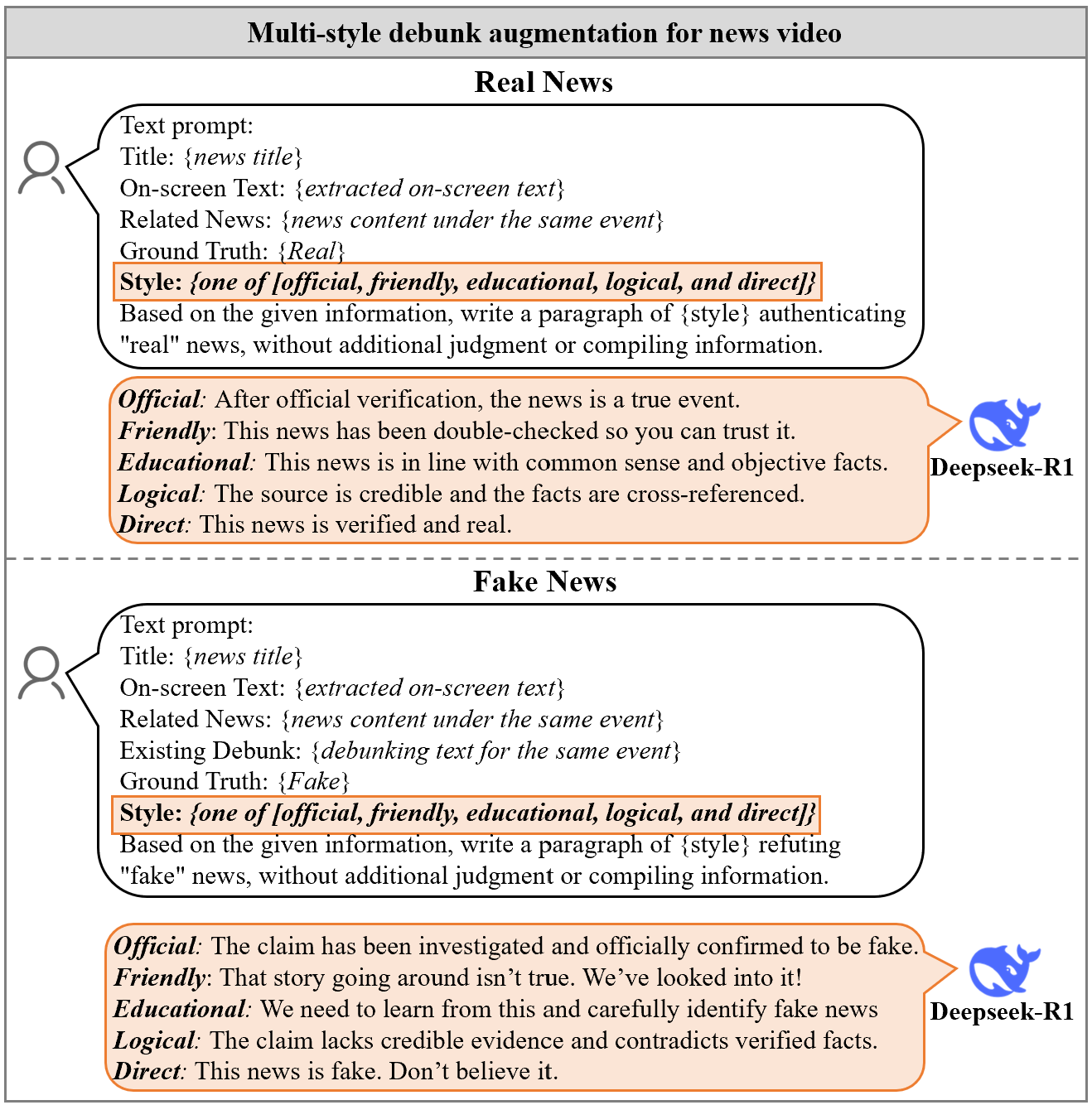}
\vspace{-1.0em}
	\caption{The pipeline and examples of multi-style debunk augmentation.}
 \label{fig:aug_styles}
\vspace{-1.5em}
\end{figure}

Using large language models (LLMs) to generate diverse training data enhances model robustness and generalization by exposing them to varied linguistic patterns and reducing overfitting. To improve the diversity and adaptability of debunking texts, we adopt an instruction-guided data augmentation strategy. Specifically, we design five distinct prompting styles listed as official, friendly, educational, logical, and direct, each promoting a unique rhetorical approach to generate various but semantically consistent debunking texts. multi-style debunk augmentation pipeline and examples are shown in Figure~\ref{fig:aug_styles}

During training, we adopt a hybrid sampling strategy to ensure both authenticity and diversity in debunk texts. For each training instance, we prioritize the use of ground-truth debunk texts provided in the original dataset when available. To complement these, we randomly sample additional debunk texts generated by the aforementioned LLM augmentation method. This ensures that each instance is associated with a total of five debunk texts, promoting the debunk diffusion model’s training across stylistic variations while preserving alignment with human-annotated content.

\subsection{Chain of Debunk}

\begin{figure}[htbp]
	\centering
	\includegraphics[width=0.48\textwidth]{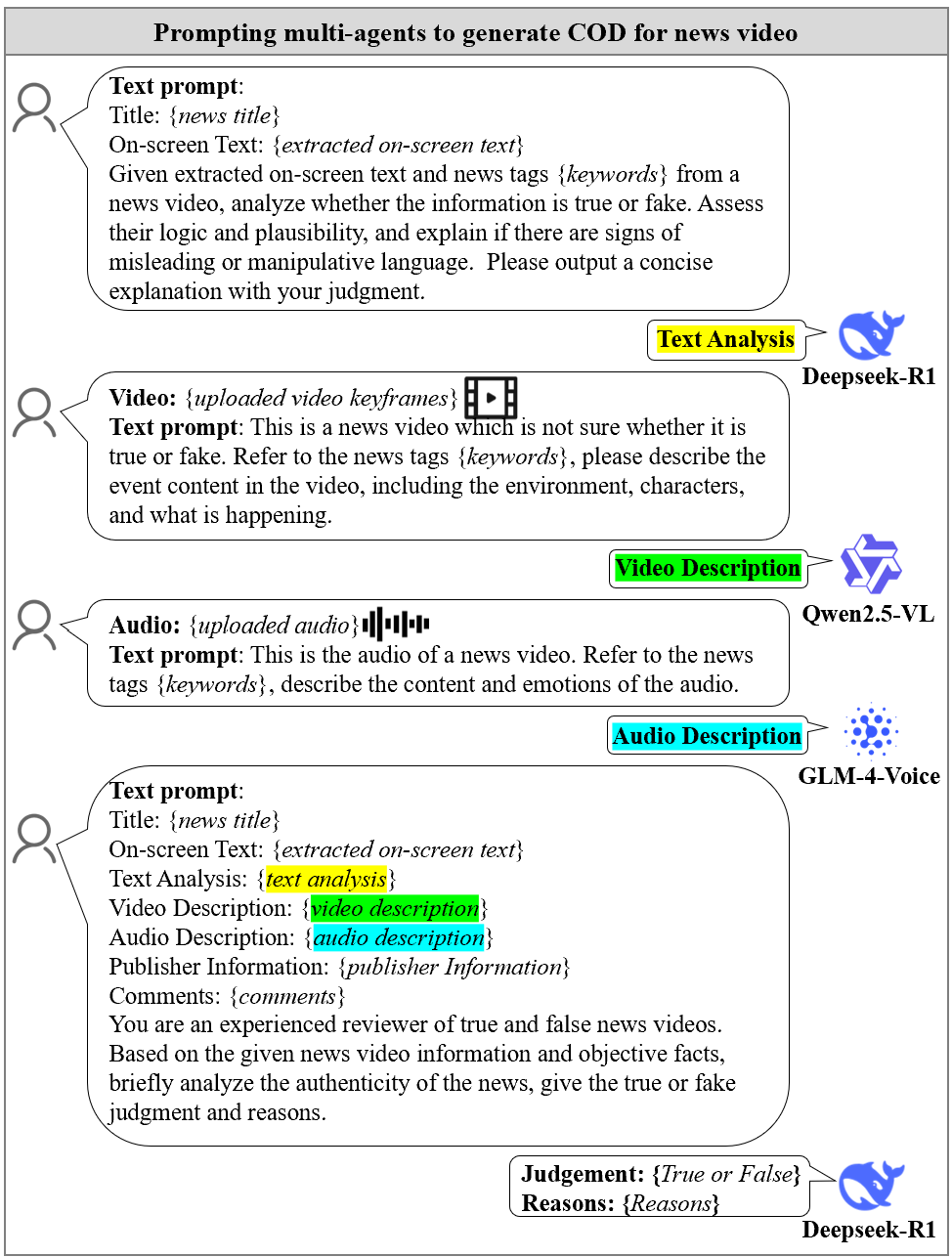}
 \vspace{-0.5em}
	\caption{The process of chain-of-debunk(COD). We use the same color to highlight related information.}
 \label{fig:cod}
\vspace{-1.5em}
\end{figure}

As shown in Figure~\ref{fig:cod}, to enhance both the accuracy and interpretability of fake news detection, we propose a multi-agent, multimodal large language model ((M)LLM) framework that extracts information from textual, visual, and audio modalities to produce a high-quality assessment, named LLM-V.

In our pipeline, the evaluation process is decomposed into a sequence of reasoning steps. Initially, dedicated agents independently process specific modalities—such as on-screen text, key video frames, and audio segments—to generate a coherent textual description of the video. These refined representations serve as the foundation for subsequent reasoning, where the system evaluates cross-modal consistency, source credibility, and factual accuracy. By structuring the reasoning process around the agents' caption analyses, the framework enables more robust and interpretable judgments regarding the authenticity of news content.

\subsubsection{Multimodal Data Understanding and Captioning}
Multimodal large language models (MLLMs) are incorporated into our framework to enrich feature representations and enable contextual understanding of video content. Leveraging their advanced language capabilities, MLLMs generate coherent descriptions from text, audio, and video, supporting deeper semantic and emotional analysis.

For textual modality, we employ the large language model Deepseek-R1, denoted as $\Phi$, to conduct text analysis on its title and the on-screen text, combining objective facts and common sense.
Following this, we leverage multimodal large language models (MLLMs) to process both keyframes and audio streams from videos, generating captions that effectively summarize the visual and auditory content. Specifically, we apply the vision-language model (VLM) Qwen2.5-VL, denoted as $\Phi_v$, to produce semantic captions for keyframes extracted from the video, providing a structured and context-aware representation of the visual content. Simultaneously, we utilize the audio-language model (ALM) GLM-4-Voice, denoted as $\Phi_a$, to generate audio captions that capture both the semantic meaning and emotional tone of the speech or sound. By integrating the news video extracted on-screen text, text analysis, visual, auditory descriptions and extra information such as comments or author profiles, we construct a comprehensive multimodal representation of the news video content. What's more, to prevent significant semantic deviations, we utilize Spacy\footnote{\url{https://github.com/explosion/spaCy}} to perform entity segmentation on event titles, extracting key entities and relevant objects, donated as keywords $\mathcal{E}$. Based on these keywords, we guide the LLM in generating the above outputs that remain closely aligned with the core subject matter, thereby preserving the original context and preventing misinterpretations or distortions. Thus, the comprehensive multimodal understanding and captioning process can be expressed as:
\begin{align}
    \mathcal{I}_t = \Phi(\mathcal{T},\mathcal{E},\mathcal{P}_t) \\
    \mathcal{I}_v = \Phi_v(\mathcal{V},\mathcal{E},\mathcal{P}_v) \\
    \mathcal{I}_a = \Phi_a(\mathcal{A},\mathcal{E},\mathcal{P}_a)
\end{align}
where $\mathcal{I}_t,\mathcal{I}_v,\mathcal{I}_a$ represent the outputted text analysis, vision caption and audio caption. $\mathcal{T},\mathcal{V},\mathcal{A}$ are the original text, vision, and audio inputs, and $\mathcal{P}_t,\mathcal{P}_v,\mathcal{P}_a$ are the corresponding suffix prompts for specific generation.

\subsubsection{LLM for News Verification}\label{LLM Verification}
At this stage, we utilize an advanced large language model (LLM), e.g., Deepseek-R1, to conduct veracity analysis by integrating textual data, linguistic features, and the audio/video captions generated by the MLLMs described earlier. The LLM assesses linguistic coherence, sentiment alignment, factual consistency, and cross-modal correlations to derive a veracity judgment. Furthermore, the system records both the classification results and the underlying rationale for each decision, which enhances transparency and provides informative cues for downstream model fusion. The process can be formulated as:
\begin{align}
    \{\mathcal{A},\mathcal{R}\}=\Phi(\mathcal{T},\mathcal{I}_t,\mathcal{I}_v,\mathcal{I}_a,\mathcal{E},\mathcal{P}),
\end{align}
where $\mathcal{T}, \mathcal{I}_t,\mathcal{I}_v,\mathcal{I}_a,\mathcal{E},\mathcal{P}_v$ refer to news textual information, text analysis, vision caption, audio caption, event entities and suffix prompt for verification, $\mathcal{A}$ is the judgment answer and $\mathcal{R}$ represents reason contents. To mitigate the inherent randomness in large language models (LLMs) and obtain more diverse and reliable outputs, we adopt a strategy of repeated inference when analyzing the veracity of news content. Specifically, each news sample is processed three times independently by the LLM, and its judgments and reasoning contents are recorded and encoded by BGE to get the chain-of-debunk feature $\mathcal{F}_{c}$.

\subsection{Veri-Verdict Fusion}
In order to integrate the textual information from multiple sources and multimodal features, we introduce a dual fusion technique named Veri-Verdict Fusion. 
In VeriFusion, we employ attention mechanisms to fuse extracted features, divided into textual-debunk and multimodal branches named TD-Fusion and MM-Fusion. In Verdict Fusion, the decisions from two branchs are fused for final judgement.

In the TD-Fusion, we enhance the semantic understanding of the textual content by utilizing the self-attention mechanism and gated-attention module. 
Self-attention captures internal textual consistency, while gated attention adaptively integrates original text, denoised debunking cues $\hat{\mathcal{F}_d}$ and chain-of-debunk features $\mathcal{F}_{c}$, effectively controlling the flow of external features into the final representation. These outputs are concatenated and passed through a two-layer MLP to produce the textual-debunk prediction score $\hat{Y}_{\text{td}}$.

For the MM-Fusion, we perform dual-modal fusion by combining the extracted textual representation with both the visual and auditory modalities separately. The text modality is extracted by audio-aligned and vision-aligned extractors donated as $F_{t_a}$ and $F_{t_v}$, while the audio feature is $F_a$ and the vision feature is $F_v$. Audio-text embeddings $\mathcal{H}_{at}$ and vision-text embeddings $\mathcal{H}_{vt}$ are obtained through cross-attention, which enables the model to capture the consistency or conflict between modalities, which is particularly beneficial for identifying incongruities in misinformation. These features are then concatenated and fed into a two-layer MLP to derive the multimudal predicted score$\hat{Y}_{\text{mm}}$. It can be formulated as
\begin{align}
    \mathcal{H}_{at} = \text{Cross-Attn}(\mathcal{F}_{t_a},\mathcal{F}_{a})\\
    \mathcal{H}_{vt} = \text{Cross-Attn}(\mathcal{F}_{t_v},\mathcal{F}_{v})\\
    \hat{Y}_{\text{mm}} = \text{MLP}([\mathcal{H}_{at},\mathcal{H}_{vt}])
\end{align}

To perform Verdict Fusion, we adopt a late fusion strategy to obtain the final prediction score $\hat{Y}_{FND}$ based on the prediction scores $\hat{Y}_{\text{mm}}$ from MM-usion and $\hat{Y}_{\text{td}}$ from TD-Fusion: 
\begin{align} 
\hat{Y}_{\text{FND}} = \psi(\hat{Y}_{\text{mm}}, \hat{Y}_{\text{td}}) = \hat{Y}_{\text{mm}} \cdot \text{tanh}(\hat{Y}_{\text{td}}) 
\end{align}

We use cross-entropy loss to optimize the model, resulting in three corresponding losses: $\mathcal{L}_{\text{FND}}$, $\mathcal{L}_{\text{mm}}$, and $\mathcal{L}_{\text{td}}$, associated with the predictions $\hat{Y}_{\text{FND}}$, $\hat{Y}_{\text{mm}}$, and $\hat{Y}_{\text{td}}$, respectively. To balance the training objectives, the final loss is defined as a weighted sum: 
\begin{align} 
\mathcal{L} = \mathcal{L}_{\text{FND}} + \alpha \mathcal{L}_{\text{mm}} + \beta \mathcal{L}_{\text{td}} + \gamma\mathcal{L}_{\text{diff}} 
\end{align} 
where $\alpha$, $\beta$ and $\gamma$ are hyperparameters that regulate the influence of each component during backpropagation.

\begin{table*}[htbp]
\vspace{-2.0em}
\centering
\caption{Performance comparison between DIFND and baselines on the Fakesv and FVC datasets.}
\vspace{-1.0em}
\scalebox{1.2}{
\begin{threeparttable}
\begin{tabular}{c|cccccccc}
\hline
\multirow{3}{*}{Methods} & \multicolumn{8}{c}{Dataset}                                                            \\ \cline{2-9} 
                         & \multicolumn{4}{c|}{Fakesv}                        & \multicolumn{4}{c}{FVC}           \\ \cline{2-9} 
                         & Acc.  & F1    & Rec.  & \multicolumn{1}{c|}{Prec.} & Acc.   & F1     & Rec.   & Prec.  \\ \hline
GPT-4V                   & 69.15 & 69.14 & 70.93 & \multicolumn{1}{c|}{71.17} & 79.51  & 79.69  & 70.19  & \textbf{92.12}  \\
Qwen2.5-VL               & 71.77 & 72.60 & 74.86 & \multicolumn{1}{c|}{70.46} & 75.10  & 85.63  & 91.96  & 80.25  \\ 
Qwen2.5-Omni             & 69.95 & 61.75 & 48.56 & \multicolumn{1}{c|}{84.76} & 79.75  & 80.12  & 69.78  & 89.69  \\ \hdashline
TikTec                   & 75.07 & 75.04 & 75.07 & \multicolumn{1}{c|}{75.18} & 77.02  & 73.95  & 73.67  & 74.24  \\
FANVN                    & 75.04 & 75.02 & 75.04 & \multicolumn{1}{c|}{75.11} & 85.81  & 85.32  & 85.20  & 85.44  \\
SV-FEND                  & 81.05 & 81.02 & 81.05 & \multicolumn{1}{c|}{81.24} & 84.71  & 85.37  & 86.53  & 84.25  \\
MMAD                     & 82.64 & 82.63 & 82.73 & \multicolumn{1}{c|}{82.63} & 89.28  & 90.36  & 90.46  & 90.27  \\
MMSFD                    & 81.83 & 81.81 & 82.02 & \multicolumn{1}{c|}{81.81} & -      & -      & -      & -      \\
Fakingrecipe             & 85.35 & 84.83 & 84.29 & \multicolumn{1}{c|}{85.84} & 85.60* & 85.07* & 85.45* & 85.86* \\
DIFND(Ours)              & \textbf{87.68} & \textbf{87.34} & \textbf{87.02} & \multicolumn{1}{c|}{\textbf{87.87}} & \textbf{92.16}  & \textbf{92.01}  & \textbf{92.21}  & 91.88  \\ \hline
\end{tabular}
\begin{tablenotes}
\tiny
\item The best performance in each column is bolded. \\The results marked with $*$ were reproduced with a portion of the models due to the absence of on-screen text in the dataset.
\end{tablenotes}
\end{threeparttable}
}
\label{tab:baselines}
\vspace{-2.0em}
\end{table*}

\section{Experiments}
\subsection{Experimental Setup}
\subsubsection{Dataset.} 
We do experiments on two fake news video dataset Fakesv \cite{qi2023fakesv} and FVC \cite{papadopoulou2019corpus}:
\begin{itemize}[leftmargin=*]
    \item Fakesv: Fakesv is the Chinese fake news short video dataset, containing 1,827 fake, 1,827 real and 1,884 debunked videos, including multimodal information such as text, audio, images, comments, and user profiles.    
    
    \item FVC: FVC dataset comprises multilingual videos from multimedia platforms with textual news content, publisher profiles and comments attached, totally containing 727 fake, 483 real and 40 debunked videos available.
\end{itemize}

\subsubsection{Baselines.}
We selected the multimodal methods and multimodal large language models for fake video detection as baselines for comparison experiments, including TikTec\cite{shang2021multimodal}, FANVN\cite{choi2021using}, SV-FEND\cite{qi2023fakesv}, MMAD\cite{zeng2024mitigating}, MMSFD\cite{ren2024mmsfd}, Fakingrecipe\cite{bu2024fakingrecipe}, and MLLMs GPT-4V\cite{yang2023dawn}, Qwen2.5-VL\cite{bai2025qwen2}, and Qwen2.5-Omni\cite{xu2025qwen2}.

\subsubsection{Implementation Details.} 
In our multimodal authenticity analysis phase, we utilize the Qwen2.5-VL-7B-Instruct\cite{bai2025qwen2} to extract visual captions from keyframes and GLM-4-Voice\cite{zeng2024glm} for audio captioning. Deepseek-R1\cite{guo2025deepseek} is employed for text-related analysis and verification tasks. 
For feature extraction, we employ BGE\cite{chen2024bge} to obtain text semantic features, while XLM-RoBERTA\cite{conneau2019unsupervised} and CLIP\cite{radford2021learning} are utilized to align text with audio and visual features, respectively. Audio features $F_a$ are extracted using CLAP\cite{wu2023large} and visual features $\mathcal{F}_v$ are derived through CLIP\cite{radford2021learning}.

For the Fakesv dataset, we split it into training, validation, and test sets in chronological order with a ratio of 70\%:15\%:15\%. For the FVC dataset, we use five-fold cross-validation and ensure that there is no event overlap among different sets. We use bge-large-zh-v1.5\footnote{\url{https://huggingface.co/BAAI/bge-large-zh-v1.5}} for Fakesv and bge-m3\footnote{\url{https://huggingface.co/BAAI/bge-m3}} for FVC to obtain textual semantic embeddings. The model is optimized using the Adam optimizer with a learning rate of 0.0001, weight decay of 5e-3, and batch size of 64, with the debunk diffusion process warmed up for 20 epochs first. Evaluation is conducted using accuracy, F1-score, recall, and precision.

\subsection{Main Results}
The performance comparison between DIFND and baselines is shown in Table~\ref{tab:baselines}, and we can draw the following conclusions.

We implement zero-shot approaches based on multimodal large language (MLLMs), but they demonstrate inferior performance compared to methods specifically optimized for fake news video detection, highlighting the inherent complexity of the fake news detection task. Furthermore, MLLMs always exhibit biases in their authenticity assessments, leading to significant risks of false alarms or missed detections. This is evident from the abnormal imbalance observed between recall and precision, which further underscores the challenges faced by these models in this domain.

As for neural network-based baselines, the comparison experiments demonstrate that DIFND outperforms all competing methods, achieving 2.33\% and 2.88\% on Fakesv and FVC datasets respectively, reflecting more reasonable information integration and logical inference of our proposed framework.

\subsection{Ablation Studies}

\subsubsection{Effects of different components}
We conduct comprehensive ablation studies to evaluate the contribution of each component in our framework, as presented in Table~\ref{tab:modules}. We begin by examining the performance of three core modules: MM-Fusion, Debunk Diffusion, and chain-of-debunk. The results show that the chain-of-debunk exhibits noticeable bias, primarily due to the inherent judgment tendencies of LLMs. Nevertheless, it provides valuable reasoning content, contributing to a 1.89\% improvement in accuracy when incorporated into the model, highlighting a trade-off between bias and the utility of model-generated explanations. In addition, Debunk Diffusion generates effective debunking features that support news verification, resulting in a further 1.69\% gain in accuracy. These modules complement each other well, and their integration leads to a total improvement of 3.34\% in accuracy compared to the base Multimodal Fusion model.

\begin{table*}[htbp]
\vspace{-2.0em} 
\centering
\caption{Ablation experiment of different model components on the Fakesv and FVC datasets. \\(MM-F: MM-Fusion, DD: Debunk Diffusion, COD: chain-of-debunk)}
\vspace{-1.0em} 
\scalebox{1.2}{
\begin{threeparttable}
\begin{tabular}{ccc|cccccccc}
\hline
\multicolumn{3}{c|}{Module}                                                                                                     & \multicolumn{8}{c}{Dataset}                                                                                                                                \\ \hline
\multirow{2}{*}{MM-F} & \multirow{2}{*}{\begin{tabular}[c]{@{}c@{}}DD\end{tabular}} & \multirow{2}{*}{COD} & \multicolumn{4}{c|}{Fakesv}                                                            & \multicolumn{4}{c}{FVC}                                           \\ \cline{4-11} 
                           &                                                                             &                      & Acc.           & F1             & Rec.           & \multicolumn{1}{c|}{Prec.}          & Acc.           & F1             & Rec.           & Prec.          \\ \hline
                           &                                                                             & \checkmark                     & 75.70          & 70.93          & 59.37          & \multicolumn{1}{c|}{\textbf{88.08}} & 87.51          & 90.75          & \textbf{94.81} & 87.95          \\
                           & \checkmark                                                                            &                      & 81.71          & 81.49          & 81.69          & \multicolumn{1}{c|}{81.38}          & 89.55          & 89.03          & 88.85          & 89.58          \\
\checkmark                           &                                                                             &                      & 84.32          & 83.50          & 82.99          & \multicolumn{1}{c|}{84.81}          & 87.64          & 86.54          & 86.18          & 88.58          \\
\checkmark                           & \checkmark                                                                            &                      & 86.01          & 85.39          & 85.39          & \multicolumn{1}{c|}{86.04}          & 91.13          & 90.65          & 90.49          & 91.02          \\
\checkmark                           &                                                                             & \checkmark                     & 86.21          & 85.22          & 85.22          & \multicolumn{1}{c|}{84.95}          & 90.34          & 89.83          & 89.67          & 90.16          \\
\checkmark                           & \checkmark                                                                            & \checkmark                     & \textbf{87.68} & \textbf{87.34} & \textbf{87.02} & \multicolumn{1}{c|}{87.87}          & \textbf{92.16} & \textbf{92.01} & 92.21          & \textbf{91.88} \\ \hline
\end{tabular}
\begin{tablenotes}
\tiny
\item The best performance in each column is bolded.
\end{tablenotes}
\end{threeparttable}
}
\vspace{-1.5em} 
\label{tab:modules}
\end{table*}

\subsubsection{Cooperation of different modalities}
To further investigate the contribution and interaction of different modalities within individual components of our framework, we conduct a series of modality-specific ablation experiments, and performance is measured by accuracy and summarized in Table~\ref{Tab:modal}. The results indicate that visual information provides more significant benefits than audio features, yielding a greater improvement in accuracy when combined with textual inputs. This suggests that visual cues are more complementary to language-based reasoning in our task setting. Notably, the best performance is achieved when all three modalities-text, vision, and audio-are fused, demonstrating the synergistic potential of multimodal integration. These findings highlight the importance of leveraging rich, heterogeneous sources of information to enhance the robustness and accuracy of misinformation detection and reasoning modules.

\begin{table}[htbp]
\vspace{-1.5em} 
\centering
\caption{Impact of modalities on the Performance of different modules on the Fakesv datasets.}
\vspace{-1.0em}
\scalebox{1.05}{
\begin{tabular}{ccc|ccc}
\hline
\multicolumn{3}{c|}{Modality} & \multicolumn{3}{c}{Module}             \\ \hline
Text    & Vision    & Audio   & \begin{tabular}[c]{@{}c@{}}MM-Fusion\end{tabular} & \begin{tabular}[c]{@{}c@{}}Debunk\\ Diffusion\end{tabular} & \begin{tabular}[c]{@{}c@{}}COD\end{tabular} \\ \hline
\checkmark       &           &         & -                                                           & 79.10                                                      & 70.83                                                            \\
\checkmark       & \checkmark         &         & 83.58                                                       & 81.34                                                      & 74.39                                                            \\
\checkmark       &           & \checkmark       & 80.59                                                       & 79.15                                                      & 72.02                                                            \\
\checkmark       & \checkmark         & \checkmark       & 84.32                                                       & 81.71                                                      & 75.70     \\ \hline                                                     
\end{tabular}
}
\vspace{-1.0em} 
\label{Tab:modal}
\end{table}

\subsubsection{Effect of debunk data augmentation}
In our experiments, we observe from Table~\ref{Tab:debunk_augment} that leveraging large language models (LLMs) for debunk-based data augmentation substantially improves both the quality of generated samples and the overall performance of downstream models. By prompting LLMs to generate counterclaims, fact-checking explanations, or clarifying statements based on annotated instances, we are able to create enriched and semantically diverse debunk data for training. Quantitative results indicate that models trained on LLM-augmented datasets achieve higher scores on key metrics compared to models trained on the original data alone, indicating that diversified data helps improve the generalization of the model. 

\begin{table}[htbp]
\vspace{-1.5em} 
\centering
\caption{Effect of debunk data augmentation on the Fakesv datasets.\\
(DD: Debunk Diffusion, DIFND: overall framework)}
\vspace{-1.0em}
\scalebox{1.2}{
\begin{tabular}{c|cccc}
\hline
\multirow{3}{*}{Debunk Source} & \multicolumn{4}{c}{Model}                                      \\ \cline{2-5} 
                               & \multicolumn{2}{c|}{DD}            & \multicolumn{2}{c}{DIFND} \\ \cline{2-5} 
                               & Acc.  & \multicolumn{1}{c|}{F1}    & Acc.        & F1          \\ \hline
Original                       & 77.98 & \multicolumn{1}{c|}{76.37} & 85.45       & 85.11       \\
LLM-Augmented                  & 81.71 & \multicolumn{1}{c|}{81.49} & 87.68       & 87.34       \\ \hline
\end{tabular}
}
\vspace{-1.0em} 
\label{Tab:debunk_augment}
\end{table}

\subsubsection{Effect of hyperparameters}
\begin{figure}[htbp]
\vspace{-2.0em} 
	\centering
    \subfloat[Effect of hyperparameters on accuracy]{
		\includegraphics[width=0.47\linewidth]{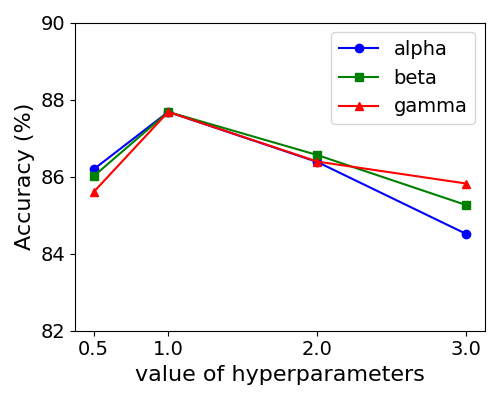}} \hspace{0.3em}
    \subfloat[Effect of hyperparameters on F1 score]{
		\includegraphics[width=0.47\linewidth]{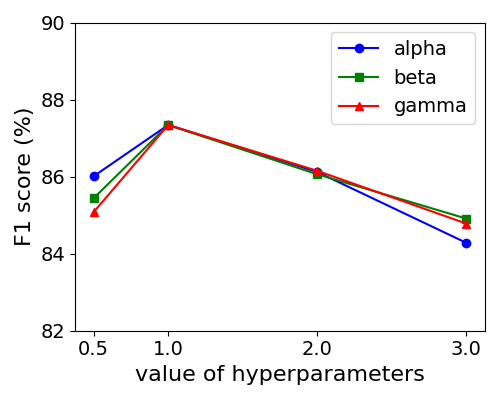}}\\
\caption{Effect of hyperparameters on the Fakesv datasets.}
\vspace{-1.0em}
\label{Fig:hyperparameters}
\end{figure}

We compared the performance of DIFND with different values
of hyperparameters $\alpha$, $\beta$ and $\gamma$ for analysis as shown in Figure~\ref{Fig:hyperparameters}. The optimal value for all three hyperparameters appears to be around 1.0, indicating the importance of balance between different training tasks. Alpha shows the steepest ascent from smaller to 1.0, demonstrating the increased importance of debunk data for performance enhancement. As the value of beta and gamma increases to greater than 1, the performance decreases faster, indicating that the imbalance between the two branches will lead to a decrease in overall performance.

\section{Conclusion}
In this paper, we presented DIFND, a novel multimodal fake news detection framework that integrates generative debunking and collaborative inference to enhance both performance and interpretability. By leveraging diffusion models to synthesize debunking or verification evidence and employing multi-agent multimodal large language for reasoning augmentation, DIFND effectively captures the complex semantics and cross-modal inconsistencies inherent in fake news content. Our unified architecture fuses raw multimodal features, generated debunking cues, and logic-driven reasoning paths to support accurate and explainable veracity assessments.
Experiments on the Fakesv and FVC datasets demonstrate that DIFND significantly outperforms existing baselines, particularly in scenarios with limited annotated debunking resources or ambiguous multimodal content. The results highlight the effectiveness of combining generation and inference as a synergistic strategy for tackling real-world misinformation.

\bibliographystyle{IEEEtran}

\begin{thebibliography}{10}
\providecommand{\url}[1]{#1}
\csname url@samestyle\endcsname
\providecommand{\newblock}{\relax}
\providecommand{\bibinfo}[2]{#2}
\providecommand{\BIBentrySTDinterwordspacing}{\spaceskip=0pt\relax}
\providecommand{\BIBentryALTinterwordstretchfactor}{4}
\providecommand{\BIBentryALTinterwordspacing}{\spaceskip=\fontdimen2\font plus
\BIBentryALTinterwordstretchfactor\fontdimen3\font minus \fontdimen4\font\relax}
\providecommand{\BIBforeignlanguage}[2]{{%
\expandafter\ifx\csname l@#1\endcsname\relax
\typeout{** WARNING: IEEEtran.bst: No hyphenation pattern has been}%
\typeout{** loaded for the language `#1'. Using the pattern for}%
\typeout{** the default language instead.}%
\else
\language=\csname l@#1\endcsname
\fi
#2}}
\providecommand{\BIBdecl}{\relax}
\BIBdecl

\bibitem{jordan2024rise}
J.~M. Jordan, \emph{The Rise of the Algorithms: How YouTube and TikTok Conquered the World}.\hskip 1em plus 0.5em minus 0.4em\relax Penn State Press, 2024.

\bibitem{shu2019beyond}
K.~Shu, S.~Wang, and H.~Liu, ``Beyond news contents: The role of social context for fake news detection,'' in \emph{Proceedings of the twelfth ACM international conference on web search and data mining}, 2019, pp. 312--320.

\bibitem{serrano2020nlp}
J.~C.~M. Serrano, O.~Papakyriakopoulos, and S.~Hegelich, ``Nlp-based feature extraction for the detection of covid-19 misinformation videos on youtube,'' in \emph{Proceedings of the 1st Workshop on NLP for COVID-19 at ACL 2020}, 2020.

\bibitem{yang2019exposing}
X.~Yang, Y.~Li, and S.~Lyu, ``Exposing deep fakes using inconsistent head poses,'' in \emph{ICASSP 2019-2019 IEEE International Conference on Acoustics, Speech and Signal Processing (ICASSP)}.\hskip 1em plus 0.5em minus 0.4em\relax IEEE, 2019, pp. 8261--8265.

\bibitem{singhal2020spotfake+}
S.~Singhal, A.~Kabra, M.~Sharma, R.~R. Shah, T.~Chakraborty, and P.~Kumaraguru, ``Spotfake+: A multimodal framework for fake news detection via transfer learning (student abstract),'' in \emph{Proceedings of the AAAI conference on artificial intelligence}, vol.~34, no.~10, 2020, pp. 13\,915--13\,916.

\bibitem{wang2022fmfn}
J.~Wang, H.~Mao, and H.~Li, ``Fmfn: Fine-grained multimodal fusion networks for fake news detection,'' \emph{Applied Sciences}, vol.~12, no.~3, p. 1093, 2022.

\bibitem{peng2024not}
L.~Peng, S.~Jian, Z.~Kan, L.~Qiao, and D.~Li, ``Not all fake news is semantically similar: Contextual semantic representation learning for multimodal fake news detection,'' \emph{Information Processing \& Management}, vol.~61, no.~1, p. 103564, 2024.

\bibitem{wei2023modeling}
L.~Wei, D.~Hu, W.~Zhou, and S.~Hu, ``Modeling both intra-and inter-modality uncertainty for multimodal fake news detection,'' \emph{IEEE Transactions on Multimedia}, 2023.

\bibitem{wang2023cross}
L.~Wang, C.~Zhang, H.~Xu, Y.~Xu, X.~Xu, and S.~Wang, ``Cross-modal contrastive learning for multimodal fake news detection,'' in \emph{Proceedings of the 31st ACM international conference on multimedia}, 2023, pp. 5696--5704.

\bibitem{qi2023fakesv}
P.~Qi, Y.~Bu, J.~Cao, W.~Ji, R.~Shui, J.~Xiao, D.~Wang, and T.-S. Chua, ``Fakesv: A multimodal benchmark with rich social context for fake news detection on short video platforms,'' in \emph{Proceedings of the AAAI Conference on Artificial Intelligence}, vol.~37, no.~12, 2023, pp. 14\,444--14\,452.

\bibitem{song2021multimodal}
C.~Song, N.~Ning, Y.~Zhang, and B.~Wu, ``A multimodal fake news detection model based on crossmodal attention residual and multichannel convolutional neural networks,'' \emph{Information Processing \& Management}, vol.~58, no.~1, p. 102437, 2021.

\bibitem{miyazaki2023fake}
K.~Miyazaki, T.~Uchiba, K.~Tanaka, J.~An, H.~Kwak, and K.~Sasahara, ``" this is fake news": Characterizing the spontaneous debunking from twitter users to covid-19 false information,'' in \emph{Proceedings of the International AAAI Conference on Web and Social Media}, vol.~17, 2023, pp. 650--661.

\bibitem{qi2023two}
P.~Qi, Y.~Zhao, Y.~Shen, W.~Ji, J.~Cao, and T.-S. Chua, ``Two heads are better than one: Improving fake news video detection by correlating with neighbors,'' \emph{arXiv preprint arXiv:2306.05241}, 2023.

\bibitem{yang2023wsdms}
R.~Yang, W.~Gao, J.~Ma, H.~Lin, and Z.~Yang, ``Wsdms: debunk fake news via weakly supervised detection of misinforming sentences with contextualized social wisdom,'' \emph{arXiv preprint arXiv:2310.16579}, 2023.

\bibitem{glockner2022missing}
M.~Glockner, Y.~Hou, and I.~Gurevych, ``Missing counter-evidence renders nlp fact-checking unrealistic for misinformation,'' \emph{arXiv preprint arXiv:2210.13865}, 2022.

\bibitem{glockner2024ambifc}
M.~Glockner, I.~Stali{\=u}nait{\.e}, J.~Thorne, G.~Vallejo, A.~Vlachos, and I.~Gurevych, ``Ambifc: Fact-checking ambiguous claims with evidence,'' \emph{Transactions of the Association for Computational Linguistics}, vol.~12, pp. 1--18, 2024.

\bibitem{chrysidis2024credible}
Z.~Chrysidis, S.-I. Papadopoulos, S.~Papadopoulos, and P.~Petrantonakis, ``Credible, unreliable or leaked?: Evidence verification for enhanced automated fact-checking,'' in \emph{Proceedings of the 3rd ACM International Workshop on Multimedia AI against Disinformation}, 2024, pp. 73--81.

\bibitem{zhou2024survey}
Y.~Zhou, C.~Guo, X.~Wang, Y.~Chang, and Y.~Wu, ``A survey on data augmentation in large model era,'' \emph{arXiv preprint arXiv:2401.15422}, 2024.

\bibitem{chai2025text}
Y.~Chai, H.~Xie, and J.~S. Qin, ``Text data augmentation for large language models: A comprehensive survey of methods, challenges, and opportunities,'' \emph{arXiv preprint arXiv:2501.18845}, 2025.

\bibitem{cheng2025empowering}
F.~Cheng, H.~Li, F.~Liu, R.~van Rooij, K.~Zhang, and Z.~Lin, ``Empowering llms with logical reasoning: A comprehensive survey,'' \emph{arXiv preprint arXiv:2502.15652}, 2025.

\bibitem{liu2024fka}
X.~Liu, P.~Li, H.~Huang, Z.~Li, X.~Cui, J.~Liang, L.~Qin, W.~Deng, and Z.~He, ``Fka-owl: Advancing multimodal fake news detection through knowledge-augmented lvlms,'' in \emph{Proceedings of the 32nd ACM International Conference on Multimedia}, 2024, pp. 10\,154--10\,163.

\bibitem{liu2024fakenewsgpt4}
------, ``Fakenewsgpt4: advancing multimodal fake news detection through knowledge-augmented lvlms,'' \emph{arXiv e-prints}, pp. arXiv--2403, 2024.

\bibitem{chalehchaleh2024enhancing}
R.~Chalehchaleh, R.~Farahbakhsh, and N.~Crespi, ``Enhancing multilingual fake news detection through llm-based data augmentation,'' in \emph{The 13th International Conference on Complex Networks and their Applications}, 2024.

\bibitem{hu2024bad}
B.~Hu, Q.~Sheng, J.~Cao, Y.~Shi, Y.~Li, D.~Wang, and P.~Qi, ``Bad actor, good advisor: Exploring the role of large language models in fake news detection,'' in \emph{Proceedings of the AAAI Conference on Artificial Intelligence}, vol.~38, no.~20, 2024, pp. 22\,105--22\,113.

\bibitem{wang2024explainable}
B.~Wang, J.~Ma, H.~Lin, Z.~Yang, R.~Yang, Y.~Tian, and Y.~Chang, ``Explainable fake news detection with large language model via defense among competing wisdom,'' in \emph{Proceedings of the ACM Web Conference 2024}, 2024, pp. 2452--2463.

\bibitem{choi2025limited}
E.~C. Choi, A.~Balasubramanian, J.~Qi, and E.~Ferrara, ``Limited effectiveness of llm-based data augmentation for covid-19 misinformation stance detection,'' \emph{arXiv preprint arXiv:2503.02328}, 2025.

\bibitem{bu2024fakingrecipe}
Y.~Bu, Q.~Sheng, J.~Cao, P.~Qi, D.~Wang, and J.~Li, ``Fakingrecipe: Detecting fake news on short video platforms from the perspective of creative process,'' in \emph{Proceedings of the 32nd ACM International Conference on Multimedia}, 2024, pp. 1351--1360.

\bibitem{liu2025modality}
Y.~Liu, Y.~Liu, Z.~Li, R.~Yao, Y.~Zhang, and D.~Wang, ``Modality interactive mixture-of-experts for fake news detection,'' \emph{arXiv preprint arXiv:2501.12431}, 2025.

\bibitem{qiu2025dsen}
Y.~Qiu, K.~Ma, W.~Zhang, R.~Pan, and Z.~Chen, ``Dsen-ek: Dual-layer semantic information extraction network with external knowledge for fake news detection,'' \emph{International Journal of Web Information Systems}, 2025.

\bibitem{jiang2025cross}
Y.~Jiang, T.~Wang, X.~Xu, Y.~Wang, X.~Song, and D.~Maynard, ``Cross-modal augmentation for few-shot multimodal fake news detection,'' \emph{Engineering Applications of Artificial Intelligence}, vol. 142, p. 109931, 2025.

\bibitem{ma2024fake}
X.~Ma, Y.~Zhang, K.~Ding, J.~Yang, J.~Wu, and H.~Fan, ``On fake news detection with llm enhanced semantics mining,'' in \emph{Proceedings of the 2024 Conference on Empirical Methods in Natural Language Processing}, 2024, pp. 508--521.

\bibitem{yi2024diffusion}
Q.~Yi, X.~Chen, C.~Zhang, Z.~Zhou, L.~Zhu, and X.~Kong, ``Diffusion models in text generation: a survey,'' \emph{PeerJ Computer Science}, vol.~10, p. e1905, 2024.

\bibitem{gong2022diffuseq}
S.~Gong, M.~Li, J.~Feng, Z.~Wu, and L.~Kong, ``Diffuseq: Sequence to sequence text generation with diffusion models,'' \emph{arXiv preprint arXiv:2210.08933}, 2022.

\bibitem{lovelace2023latent}
J.~Lovelace, V.~Kishore, C.~Wan, E.~Shekhtman, and K.~Q. Weinberger, ``Latent diffusion for language generation,'' \emph{Advances in Neural Information Processing Systems}, vol.~36, pp. 56\,998--57\,025, 2023.

\bibitem{zhou2024difflm}
Y.~Zhou, X.~Wang, Y.~Niu, Y.~Shen, L.~Tang, F.~Chen, B.~He, L.~Sun, and L.~Wen, ``Difflm: Controllable synthetic data generation via diffusion language models,'' \emph{arXiv preprint arXiv:2411.03250}, 2024.

\bibitem{chen2024effective}
Z.~Chen, L.~Wang, Y.~Wu, X.~Liao, Y.~Tian, and J.~Zhong, ``An effective deployment of diffusion lm for data augmentation in low-resource sentiment classification,'' \emph{arXiv preprint arXiv:2409.03203}, 2024.

\bibitem{wang2023incomplete}
Y.~Wang, Y.~Li, and Z.~Cui, ``Incomplete multimodality-diffused emotion recognition,'' \emph{Advances in Neural Information Processing Systems}, vol.~36, pp. 17\,117--17\,128, 2023.

\bibitem{blow2025data}
C.~H. Blow, L.~Qian, C.~Gibson, P.~Obiomon, and X.~Dong, ``Data augmentation via diffusion model to enhance ai fairness,'' \emph{Frontiers in Artificial Intelligence}, vol.~8, p. 1530397, 2025.

\bibitem{wei2022chain}
J.~Wei, X.~Wang, D.~Schuurmans, M.~Bosma, F.~Xia, E.~Chi, Q.~V. Le, D.~Zhou \emph{et~al.}, ``Chain-of-thought prompting elicits reasoning in large language models,'' \emph{Advances in neural information processing systems}, vol.~35, pp. 24\,824--24\,837, 2022.

\bibitem{yao2023react}
S.~Yao, J.~Zhao, D.~Yu, N.~Du, I.~Shafran, K.~Narasimhan, and Y.~Cao, ``React: Synergizing reasoning and acting in language models,'' in \emph{International Conference on Learning Representations (ICLR)}, 2023.

\bibitem{chen2025lvagent}
B.~Chen, Z.~Yue, S.~Chen, Z.~Wang, Y.~Liu, P.~Li, and Y.~Wang, ``Lvagent: Long video understanding by multi-round dynamical collaboration of mllm agents,'' \emph{arXiv preprint arXiv:2503.10200}, 2025.

\bibitem{shi2024unlocking}
Y.~Shi, S.~Di, Q.~Chen, and W.~Xie, ``Unlocking video-llm via agent-of-thoughts distillation,'' \emph{arXiv preprint arXiv:2412.01694}, 2024.

\bibitem{liu2024detect}
Y.~Liu, J.~Zhu, K.~Zhang, H.~Tang, Y.~Zhang, X.~Liu, Q.~Liu, and E.~Chen, ``Detect, investigate, judge and determine: A novel llm-based framework for few-shot fake news detection,'' \emph{arXiv preprint arXiv:2407.08952}, 2024.

\bibitem{zhang2025llms}
C.~Zhang, Z.~Feng, Z.~Zhang, J.~Qiang, G.~Xu, and Y.~Li, ``Is llms hallucination usable? llm-based negative reasoning for fake news detection,'' in \emph{Proceedings of the AAAI Conference on Artificial Intelligence}, vol.~39, no.~1, 2025, pp. 1031--1039.

\bibitem{papadopoulou2019corpus}
O.~Papadopoulou, M.~Zampoglou, S.~Papadopoulos, and I.~Kompatsiaris, ``A corpus of debunked and verified user-generated videos,'' \emph{Online information review}, vol.~43, no.~1, pp. 72--88, 2019.

\bibitem{shang2021multimodal}
L.~Shang, Z.~Kou, Y.~Zhang, and D.~Wang, ``A multimodal misinformation detector for covid-19 short videos on tiktok,'' in \emph{2021 IEEE international conference on big data (big data)}.\hskip 1em plus 0.5em minus 0.4em\relax IEEE, 2021, pp. 899--908.

\bibitem{choi2021using}
H.~Choi and Y.~Ko, ``Using topic modeling and adversarial neural networks for fake news video detection,'' in \emph{Proceedings of the 30th ACM international conference on information \& knowledge management}, 2021, pp. 2950--2954.

\bibitem{zeng2024mitigating}
Z.~Zeng, M.~Luo, X.~Kong, H.~Liu, H.~Guo, H.~Yang, Z.~Ma, and X.~Zhao, ``Mitigating world biases: A multimodal multi-view debiasing framework for fake news video detection,'' in \emph{Proceedings of the 32nd ACM International Conference on Multimedia}, 2024, pp. 6492--6500.

\bibitem{ren2024mmsfd}
S.~Ren, Y.~Liu, Y.~Zhu, W.~Bing, H.~Ma, and W.~Wang, ``Mmsfd: Multi-grained and multi-modal fusion for short video fake news detection,'' in \emph{2024 7th International Conference on Data Science and Information Technology (DSIT)}.\hskip 1em plus 0.5em minus 0.4em\relax IEEE, 2024, pp. 1--11.

\bibitem{yang2023dawn}
Z.~Yang, L.~Li, K.~Lin, J.~Wang, C.-C. Lin, Z.~Liu, and L.~Wang, ``The dawn of lmms: Preliminary explorations with gpt-4v (ision),'' \emph{arXiv preprint arXiv:2309.17421}, vol.~9, no.~1, p.~1, 2023.

\bibitem{bai2025qwen2}
S.~Bai, K.~Chen, X.~Liu, J.~Wang, W.~Ge, S.~Song, K.~Dang, P.~Wang, S.~Wang, J.~Tang \emph{et~al.}, ``Qwen2. 5-vl technical report,'' \emph{arXiv preprint arXiv:2502.13923}, 2025.

\bibitem{xu2025qwen2}
J.~Xu, Z.~Guo, J.~He, H.~Hu, T.~He, S.~Bai, K.~Chen, J.~Wang, Y.~Fan, K.~Dang \emph{et~al.}, ``Qwen2. 5-omni technical report,'' \emph{arXiv preprint arXiv:2503.20215}, 2025.

\bibitem{zeng2024glm}
A.~Zeng, Z.~Du, M.~Liu, K.~Wang, S.~Jiang, L.~Zhao, Y.~Dong, and J.~Tang, ``Glm-4-voice: Towards intelligent and human-like end-to-end spoken chatbot,'' \emph{arXiv preprint arXiv:2412.02612}, 2024.

\bibitem{guo2025deepseek}
D.~Guo, D.~Yang, H.~Zhang, J.~Song, R.~Zhang, R.~Xu, Q.~Zhu, S.~Ma, P.~Wang, X.~Bi \emph{et~al.}, ``Deepseek-r1: Incentivizing reasoning capability in llms via reinforcement learning,'' \emph{arXiv preprint arXiv:2501.12948}, 2025.

\bibitem{chen2024bge}
J.~Chen, S.~Xiao, P.~Zhang, K.~Luo, D.~Lian, and Z.~Liu, ``Bge m3-embedding: Multi-lingual, multi-functionality, multi-granularity text embeddings through self-knowledge distillation,'' \emph{arXiv preprint arXiv:2402.03216}, 2024.

\bibitem{conneau2019unsupervised}
A.~Conneau, K.~Khandelwal, N.~Goyal, V.~Chaudhary, G.~Wenzek, F.~Guzm{\'a}n, E.~Grave, M.~Ott, L.~Zettlemoyer, and V.~Stoyanov, ``Unsupervised cross-lingual representation learning at scale,'' \emph{arXiv preprint arXiv:1911.02116}, 2019.

\bibitem{radford2021learning}
A.~Radford, J.~W. Kim, C.~Hallacy, A.~Ramesh, G.~Goh, S.~Agarwal, G.~Sastry, A.~Askell, P.~Mishkin, J.~Clark \emph{et~al.}, ``Learning transferable visual models from natural language supervision,'' in \emph{International conference on machine learning}.\hskip 1em plus 0.5em minus 0.4em\relax PmLR, 2021, pp. 8748--8763.

\bibitem{wu2023large}
Y.~Wu, K.~Chen, T.~Zhang, Y.~Hui, T.~Berg-Kirkpatrick, and S.~Dubnov, ``Large-scale contrastive language-audio pretraining with feature fusion and keyword-to-caption augmentation,'' in \emph{ICASSP 2023-2023 IEEE International Conference on Acoustics, Speech and Signal Processing (ICASSP)}.\hskip 1em plus 0.5em minus 0.4em\relax IEEE, 2023, pp. 1--5.

\end{thebibliography}

\vspace{50em}
\appendix
\subsection{Prompts}

In this section we present the prompts used for debunk augmentation and MLLMs baselines implementation.

\subsubsection{Prompts of Debunking Text Augmentation for LLM}
\begin{figure}[htbp]
\vspace{-1em}
	\centering
	\includegraphics[width=0.45\textwidth]{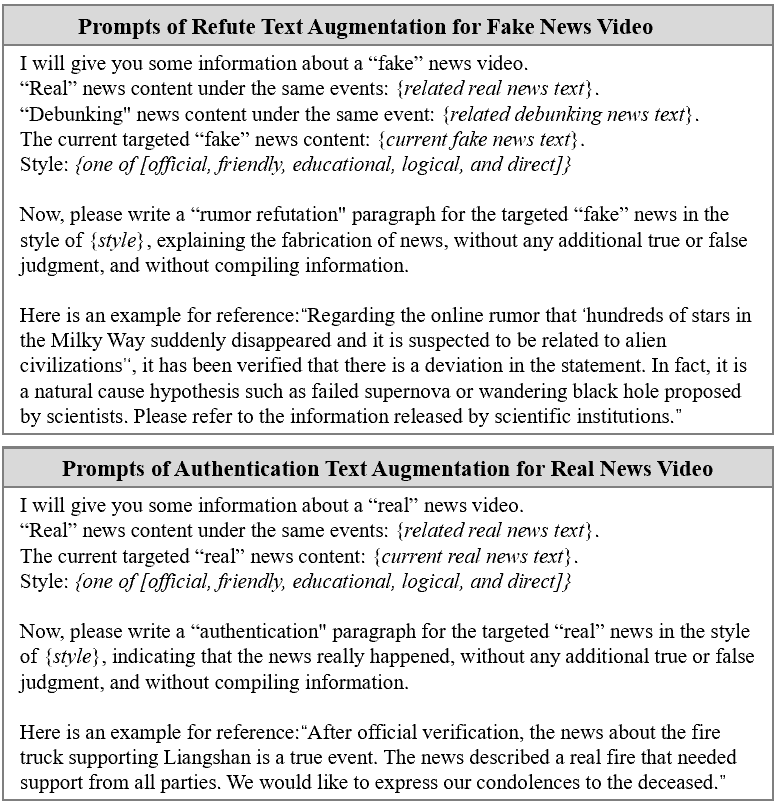}
\vspace{-1em}
\caption{Prompts of Debunking Text Augmentation for LLM.}
\label{fig:aug_prompts}
\end{figure}

As shown in Figure~\ref{fig:aug_prompts}, we design distinct prompts to augment debunking texts for fake and real news, leveraging cues from the news content itself, related news within the same event, and existing annotations.
\subsubsection{Prompts of the Detection Task for baseline MLLMs}
\begin{figure}[htbp]
\vspace{-2em}
	\centering
	\includegraphics[width=0.45\textwidth]{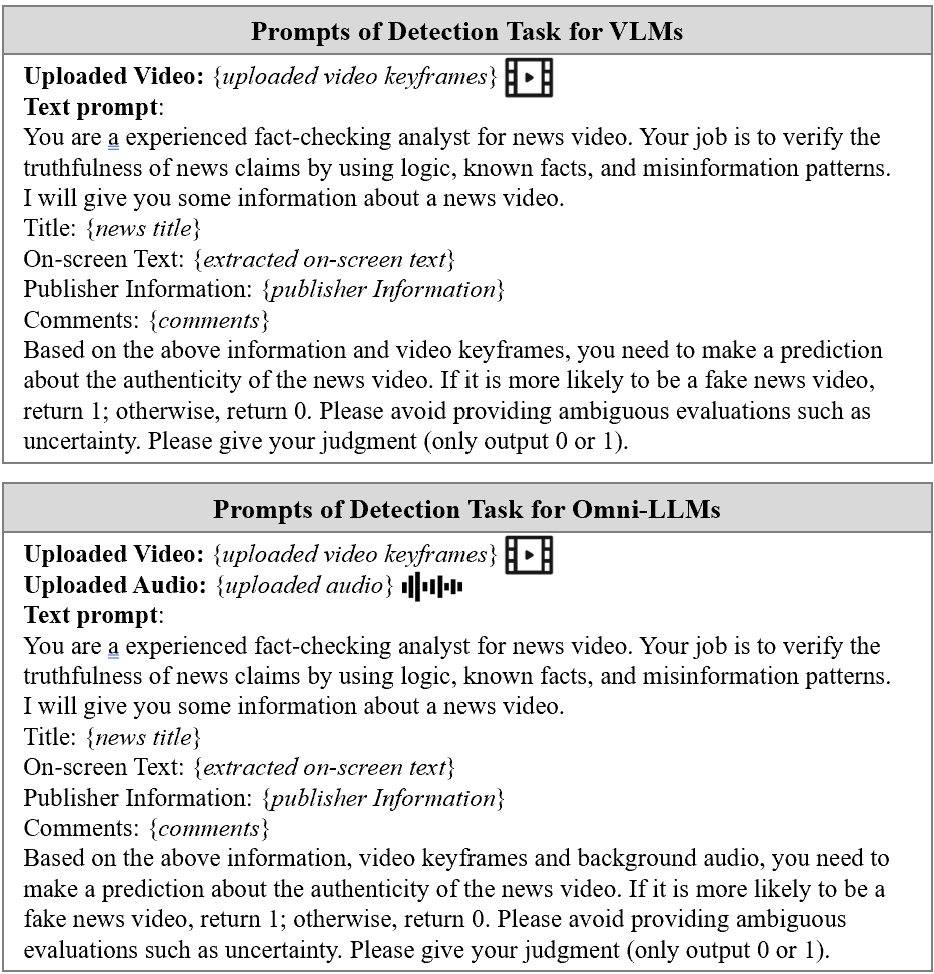}
\vspace{-1em}
\caption{Prompts of the Detection Task for baseline MLLMs.}
\label{fig:llm_prompts}
\vspace{-2em}
\end{figure}

As shown in Figure~\ref{fig:llm_prompts}, the input for vision-language models (VLMs) such as GPT-4V and Qwen2.5-VL includes video keyframes, title, on-screen text, and a textual prompt. These inputs are processed to generate the veracity judgment. For omni-modal large language models like Qwen2.5-Omni, the input is extended to include the corresponding audio sequence.

\subsection{Case Study}
We further demonstrate the capabilities of Debunk Diffusion with augmented data and chain-of-debunk in detecting fake news videos through cases from the Fakesv dataset in the following figures. The augmented debunk text not only enhances the directionality while retaining the semantic structure of the original content but also emphasizes causal relationships, evidence, and reasoning connections, thereby guiding the model to enhance the logical coherence of debunking and helping debunk Diffusion training to generate. The chain-of-debunk (COD) module detects internal contradictions (e.g., mismatches between text content and objective facts) as well as inconsistencies between modalities (e.g., a video transcript contradicting its visual evidence or audio characteristics). Three typical cases are shown as follows.




\begin{center}
  \includegraphics[width=0.49\textwidth]{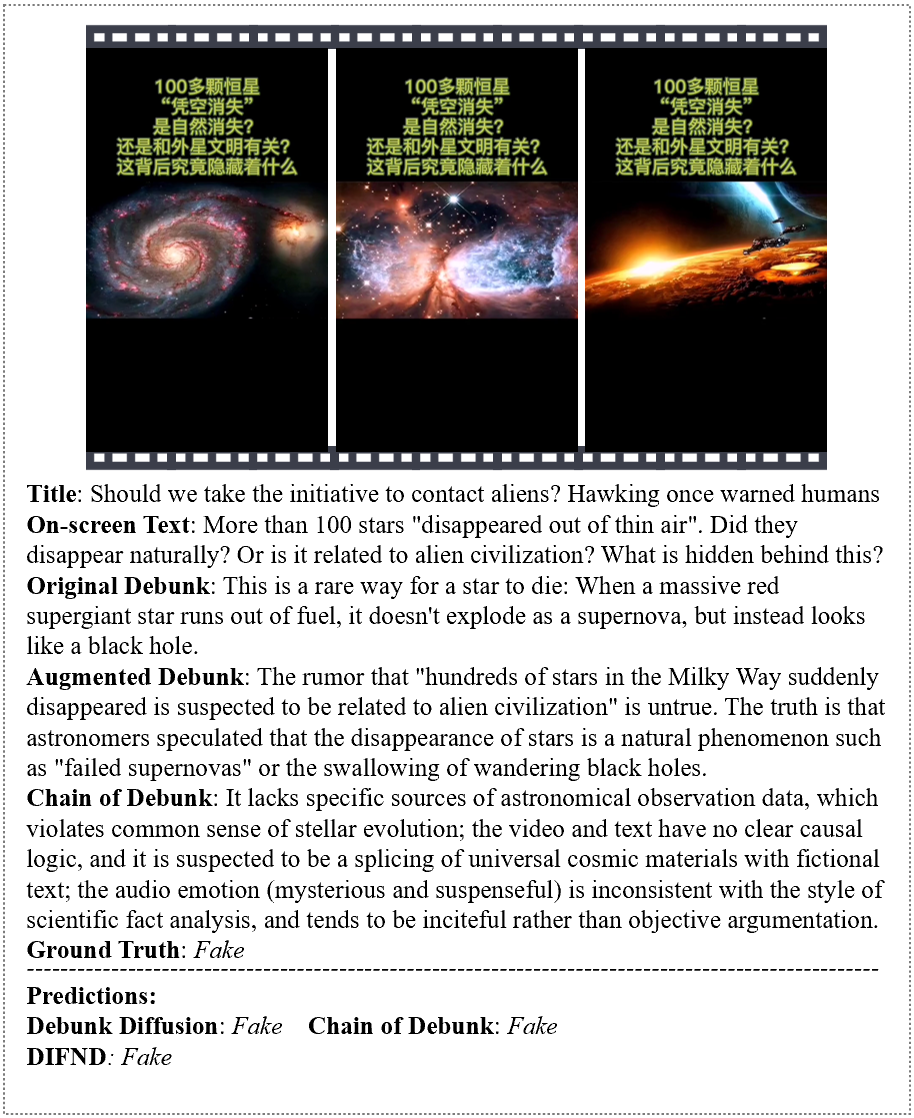}
\end{center}

\begin{figure}[htbp]
    \centering
    \includegraphics[width=\linewidth]{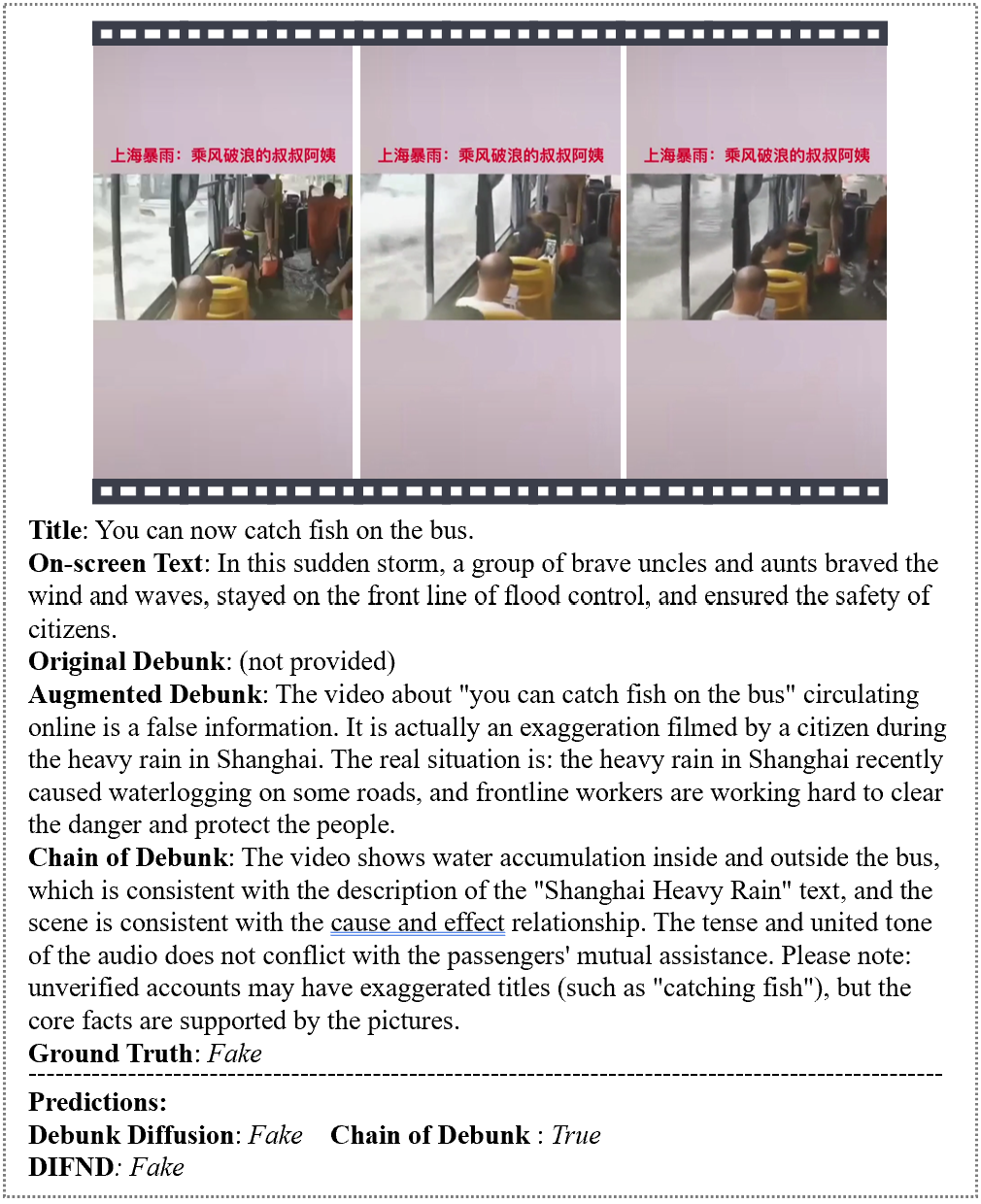}

    \vspace{0.5cm}  

    \includegraphics[width=\linewidth]{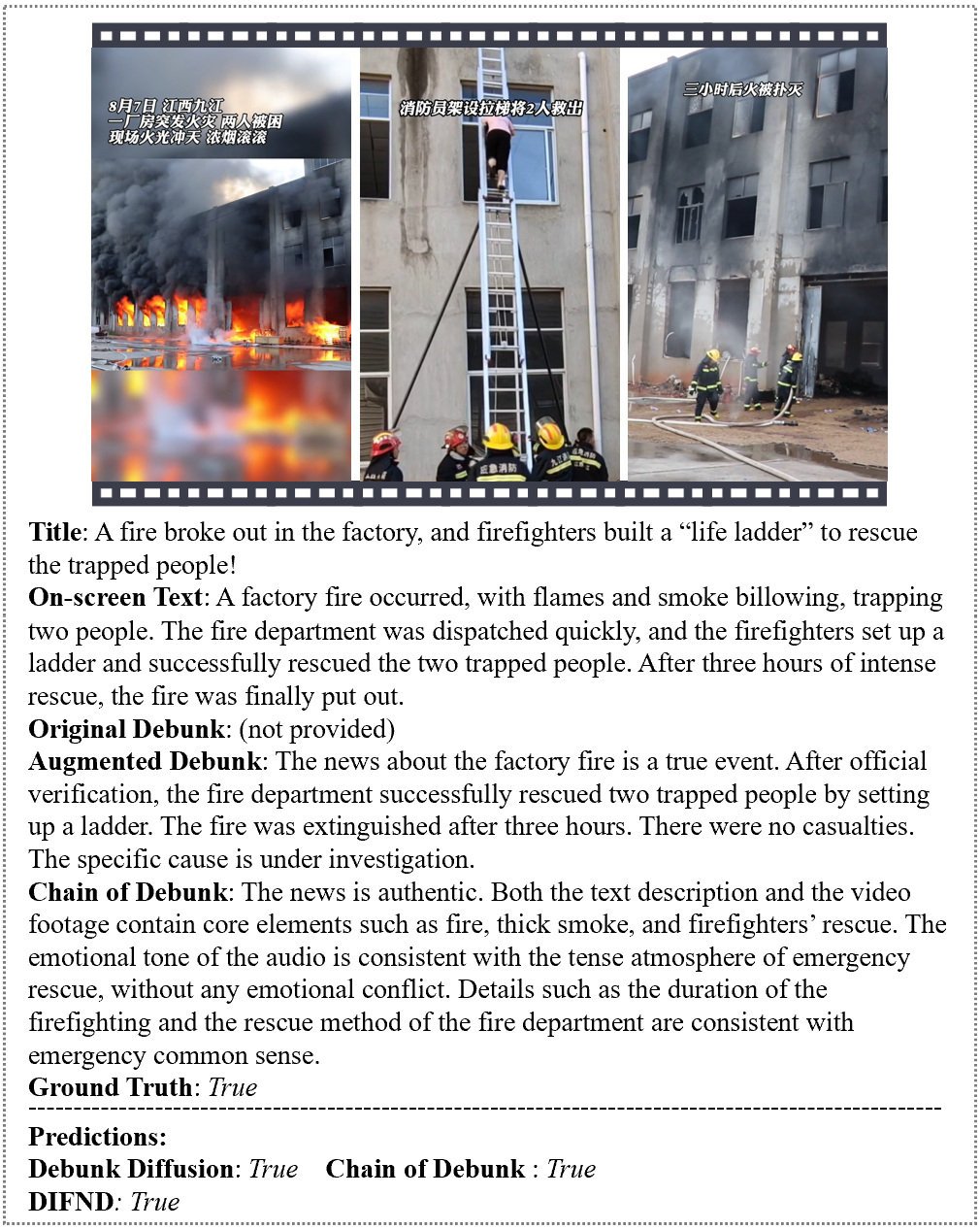}
\end{figure}

\end{document}